\documentclass{article} % For LaTeX2e
\usepackage{iclr2024_conference,times}
\usepackage{amsmath,amsfonts,bm}

\def\eqref#1{equation~\ref{#1}}
\def\1{\bm{1}}

\DeclareMathAlphabet{\mathsfit}{\encodingdefault}{\sfdefault}{m}{sl}
\SetMathAlphabet{\mathsfit}{bold}{\encodingdefault}{\sfdefault}{bx}{n}

\usepackage{hyperref}
\usepackage{url}
\usepackage{graphicx}
\usepackage{enumerate}
\usepackage{subfigure}
\usepackage{booktabs}
\usepackage{multirow}
\usepackage{amsmath}
\usepackage{amssymb}
\usepackage{subfigure}
\usepackage{wrapfig}
\usepackage{hyperref}
\usepackage{pifont}
\usepackage{color}
\usepackage[ruled]{algorithm2e}
\usepackage{algorithmic}
\usepackage{amsmath}
\usepackage{amsthm}
\usepackage{changepage}

\title{Skip \textbackslash n: A Simple Method to Reduce Hallucination in Large Vision-Language Models}
\author{Zongbo Han$^1$, Zechen Bai$^2$, Haiyang Mei$^2$, Qianli Xu$^3$, Changqing Zhang$^{1*}$, Mike Zheng Shou$^2$\thanks{Corresponding author. Work done during an internship of Zongbo Han (zongbo@tju.edu.cn) at Show Lab.}
\\
College of Intelligence and Computing, Tianjin University $^1$ \\ 
Show Lab, National University of Singapore$^2$ \\
Institute for Infocomm Research, A*STAR$^3$
}

\iclrfinalcopy % Uncomment for camera-ready version, but NOT for submission.
\begin{document}
\maketitle

\begin{abstract}
% Recent advancements in large vision-language models (LVLMs) have greatly enhanced their ability to understand visual information through human language. Despite these advances, LVLMs are susceptible to multimodal hallucinations, such as generating non-existent objects in visual contexts. The fundamental causes of these hallucinations, however, are not well understood. This paper proposes a novel perspective, identifying inherent biases in visual-language models as a significant contributor to hallucinations. Specifically, we identify a semantic shift bias related to paragraph breaks (\n\n'), where content before and after \n\n' in the training data often undergoes substantial semantic changes. This pattern leads models to infer that content following \n\n' should be markedly different from preceding content, increasing the likelihood of hallucinatory descriptions post \n\n'. Our hypothesis is validated on multiple widely-used open-source LVLMs. We demonstrate that mitigating the use of \n\n' effectively reduces hallucinations. Conversely, intentionally inserting \n\n' in descriptions can induce more hallucinations.

%Current methods to mitigate hallucinations mainly focus on model fine-tuning or employing post-processing techniques for correction. However, these methods often lack an exploration of the fundamental reasons causing hallucinations. 

\textcolor{black}{Recent advancements in large vision-language models (LVLMs) have demonstrated impressive capability in visual information understanding with human language. Despite these advances, LVLMs still face challenges with multimodal hallucination, such as generating text descriptions of objects that are not present in the visual information. However, the underlying fundamental reasons of multimodal hallucinations remain poorly explored. In this paper, we propose a new perspective, suggesting that the inherent biases in LVLMs might be a key factor in hallucinations. Specifically, we systematically identify a semantic shift bias related to paragraph breaks (`\textbackslash n\textbackslash n'), where the content before and after `\textbackslash n\textbackslash n' in the training data frequently exhibit significant semantic changes. This pattern leads the model to infer that the contents following `\textbackslash n\textbackslash n' should be obviously different from the preceding contents with less hallucinatory descriptions, thereby increasing the probability of hallucinatory descriptions subsequent to the `\textbackslash n\textbackslash n'. We have validated this hypothesis on multiple publicly available LVLMs. Besides, we find that deliberately inserting `\textbackslash n\textbackslash n' at the generated description can induce more hallucinations. A simple method is proposed to effectively mitigate the hallucination of LVLMs by skipping the output of `\textbackslash n'.} 
Code is available at \href{https://github.com/hanmenghan/Skip-n}{https://github.com/hanmenghan/Skip-n}. %Finally, we performed a statistical analysis on the instruction fine-tuning data set of the large visual language model mentioned above, indicating that this bias may originate from unreasonable instruction fine-tuning data.
\end{abstract}

\section{Introduction}
\vspace{-1mm}
\begin{figure}[!hp]
% \begin{figure}[b]
  \centering
  % \begin{minipage}{0.72\textwidth}
  %   \includegraphics[width=\linewidth]{fig.pdf}
  %   \label{fig_challenging_sample}
  % \end{minipage}\hfill
  \begin{minipage}{1\textwidth}
    \includegraphics[width=\linewidth]{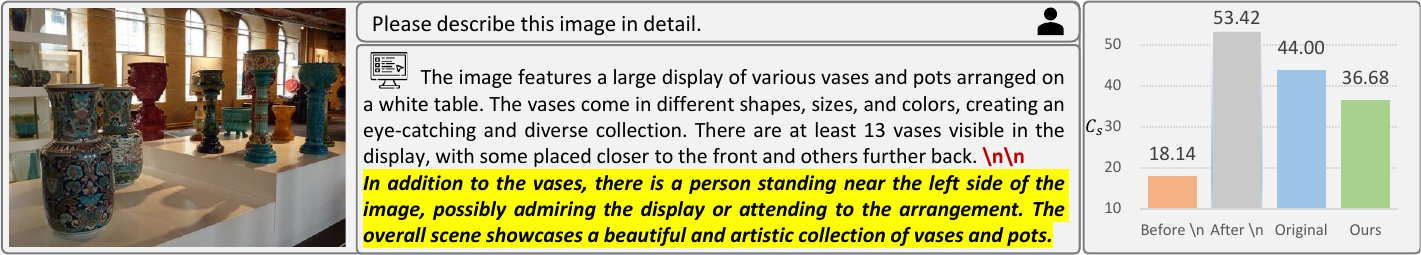}
    % \caption{Caption for the second figure.}
  \end{minipage}
  \vspace{-2mm}
 \caption{(Left) A hallucination example where the LVLM generates the hallucinatory description after the paragraph breaks `\textbackslash n\textbackslash n'. (Right) The severity of hallucinations across different outputs with BakLLaVA \citep{liu2023improved}, including Before \textbackslash n, After \textbackslash n, Original output, and Our mitigation results. Higher values indicate more severe hallucinations.} 
\label{fig_challenging_sample}
\end{figure}
% \vspace{-1mm}
% \begin{figure}[!hp]
%   \centering
%   \includegraphics[width=0.95\textwidth]{fig.pdf}
%   % \caption{Example of hallucination \q{triggered by paragraph breaks (`\textbackslash n\textbackslash n')}. As shown in the figure, after the generation of `\textbackslash n\textbackslash n', the corresponding text description begins to exhibit hallucinations.}
%   \caption{A hallucination example where the LVLM (LLaVA-v1.5-13B \citep{liu2023improved}) generates the hallucinatory description after the paragraph breaks `\textbackslash n\textbackslash n'.}
%   %TOD 放在第一页
% \label{fig_challenging_sample}
% \end{figure}

\textcolor{black}{LVLMs have demonstrated remarkable capabilities in describing and analyzing the provided visual information using human language, marking a significant step towards general artificial intelligence \citep{achiam2023gpt,zhu2023minigpt,li2023blip,liu2023improved,liu2023visual}. However, LVLMs often suffer from multimodal hallucinations, such as object hallucination where non-existent objects in the visual information are described in the generated responses \citep{wang2023evaluation, huang2023opera, zhou2023analyzing, cui2023holistic}. Such misleading responses can limit the deployment of LVLMs in many safety-critical applications such as autonomous driving~\citep{bojarski2016end} and machine-learning-aided medical diagnosis~\citep{esteva2017dermatologist}.}

Many approaches have been proposed for mitigating hallucinations in LVLMs. These methods are primarily categorized into two types, including retraining-based and post-hoc processing-based approaches. Retraining-based approaches include redesigning the vision encoder \citep{tong2024eyes}, collecting high-quality data for finetuning \citep{wang2023vigc}, and employing reinforcement learning for LVLM fine-tuning \citep{zhao2023beyond,sun2023aligning,yu2023rlhf}. Besides, post-hoc processing-based methods involve designing decoding strategies \citep{huang2023opera,leng2023mitigating} and training an additional reviser model to detect and mitigate hallucinations \citep{zhou2023analyzing}. Even though these approaches are effective in some circumstances, they still lack an exploration of key factors in LVLM hallucinations. Recent studies have found that imperfections in the vision encoder \citep{tong2024eyes} and inherent uncertainties in the models \citep{zhou2023analyzing} can lead to hallucinations. In contrast, our work investigates the occurrence of hallucinations in LVLMs from the perspective of inherent biases within the models. 

As shown in Fig.~\ref{fig_challenging_sample}, we identify a special semantic shift bias triggered by paragraph breaks, where training data often show significant semantic changes before and after `\textbackslash n\textbackslash n'. This leads to a tendency for LVLMs to deviate from the previous non-hallucinatory description after `\textbackslash n\textbackslash n', resulting in hallucinations. We validate this hypothesis across several LVLMs \citep{liu2023visual,liu2023improved,li2023blip,zhu2023minigpt,Fuyu-8B}. Besides, we explore the use of `\textbackslash n\textbackslash n' as a method to induce hallucinations in existing LVLMs. We find that inserting `\textbackslash n\textbackslash n' in generated sentences significantly increases the probability of hallucinations, which further supports our findings that `\textbackslash n\textbackslash n' increases the probability of hallucinations. Based on this observation, we propose two simple yet effective methods to reduce hallucinations, including changing the prompt on the input side and modifying logits on the output side, both aimed at avoiding the output of `\textbackslash n\textbackslash n'. Experimental results show that the proposed method significantly reduces the occurrence of hallucinations. 
% Additionally, we explore the use of `\textbackslash n\textbackslash n' as a method to induce hallucinations in existing LVLMs. We find that inserting `\textbackslash n\textbackslash n' in generated sentences significantly increases the probability of hallucinations, 
% which further supports our hypothesis that `\textbackslash n\textbackslash n' may lead to hallucinations. which further supports our findings that `\textbackslash n\textbackslash n' increases the probability of hallucinations.

Overall, our main contributions are as follows. Firstly, we identify that the inherent bias in LVLMs may be a key factor leading to hallucinations. Secondly, we discover a method to induce multimodal hallucinations, serving as an effective attack mechanism. Finally, we propose two effective and efficient solutions to reduce hallucinations in LVLMs without requiring additional costs. 
% \mhy{how about we change the order of 2nd and 3rd contributions?}

% \begin{figure}[tp]
%   \centering
%   \includegraphics[width=1\textwidth]{figure2.pdf}
%   \caption{Examples of hallucinations caused by paragraph breaks (`\textbackslash n\textbackslash n'). As shown in the figure, after generating `\textbackslash n\textbackslash n', the corresponding text description begins to exhibit hallucinations.}
%     \label{fig_challenging_sample}
% \end{figure}

% \section{Related works}
% 深度学习中的偏见

%多模态幻觉

\section{Method}
% \textbf{Basic settings}. Given the input vision information and the corresponding prompt, LVLMs typically generate sentences by incrementally predicting the next token probability based on the previous tokens. Formally, suppose that the generated sequence $S$ is composed of $N$ tokens $z$, i.e., $S=\{z_1, z_2, \cdots, z_N\}$. Given an image $x$ and the prompt $p$, the predicted logits of the next token $z_i$ at the $i$-th position in the sequence can be computed with $L_i(z_i|S_{<i},x,p)$, where $S_{<i}$ denote the sequence $S=\{z_n\}_{n=1}^{i-1}$. Based on the predicted logits $L_i(z_i|S_{<i},x,p)$, the $i$-th token probability $P_i$ can be obtained with $SoftMax (L_i)$. Then various decoding strategies can be employed to determine the next specific token. A typical decoding strategy is the greedy decoding, which selects the token with the highest predicted probability as the final prediction.

% As outlined in our introduction, our proposed method aims to reduce hallucinations in language model outputs by preventing the generation of paragraph breaks (`\textbackslash n\textbackslash n'). This approach helps to mitigate semantic shift bias in models, which can cause Large Vocabulary Language Models (LVLMs) to stray from the focus of previous descriptions, leading to hallucinatory content in the output. Our goal can be effectively achieved through two distinct yet complementary methods: modifying the prompts fed into the LVLMs and altering decoding strategies during output generation.

Our proposed method aims to reduce hallucinations by preventing the model from generating paragraph breaks (`\textbackslash n\textbackslash n'). Therefore, we can mitigate the semantic shift bias in LVLMs, which may cause the description to stray from the initial focus, leading to hallucinatory content in the following outputs. This goal can be efficiently achieved through two orthogonal methods: modifying the prompt given to the LVLMs during input and changing the decoding strategies during output. 
%todo without \n 

\textbf{Mitigating Hallucinations during Input (MiHI).} 
% From the perspective of modifying the input prompt, 
LVLMs can usually follow human instructions well through instruction tuning. Therefore, we try to modify the prompt for LVLMs and encourage them to fulfill the original instructions while avoiding the output of `\textbackslash n', thereby maintaining the continuity and coherence of the generated text. Specifically, taking the task of describing an input image as an example, the commonly used prompt is \emph{``Please describe this image in detail"}. The proposed method modify the above prompt to \emph{``Please describe this image in detail \textcolor{brown}{in one paragraph}.''} This modification emphasizes the generation of a single, continuous paragraph, thereby avoiding the output of paragraph breaks `\textbackslash n\textbackslash n'. Note that the prompt (\textcolor{brown}{\emph{in one paragraph}}) provided above can be adjusted as needed based on practical performance.

\textbf{Mitigating Hallucinations during Output (MiHO).} From the perspective of modifying the output decoding strategies, we can avoid the output of `\textbackslash n' by reducing the logits corresponding to the `\textbackslash n' token. Formally, considering the next token logits is $L$. We can adjust the next token logits to avoid outputting `\textbackslash n'. Specifically, the adjusted  next token logits $\hat{L}$ can be obtained with $\hat{L} = L-\lambda\cdot\mathbf{1}_{\backslash\textbackslash},$
where $\lambda$ is a hyperparameter used to control the penalty strength. $\mathbf{1}_{\backslash\textbackslash}$ represents a one-hot encoding vector in which the dimension corresponding to the `\textbackslash n' token is set to $1$ while all other dimensions are set to $0$. In our implementation, we set $\lambda$ to positive infinity to effectively eliminate the prediction probability of `\textbackslash n' token. In contrast, when we want to intentionally insert `\textbackslash n' to attack the model\footnote{Unlike traditional attacks, the attack described here aims to better observe the impact of the `\textbackslash n' token.}, we can adjust $\lambda$ at the specific position accordingly .
% \vspace{-4mm}
\section{Experiments}
We conduct extensive experiments on multiple LVLMs to address the following questions. Q1 Hypothesis verification: Does the description generated after `\textbackslash n' exhibit more serious hallucinations? Q2 Attackability: Does the insertion of `\textbackslash n\textbackslash n' in the generated description trigger hallucinations? Q3 Effectiveness: Can our proposed method (MiHO and MiHI) effectively mitigate hallucinations? 

\textbf{Experimental settings.} Our evaluation are conducted on the six publicly available LVLMs, including BakLLaVA, LLaVA-v1.5-7B, LLaVA-v1.5-13B \citep{liu2023improved}, InstructBLIP-7B \citep{li2023blip}, MiniGPT-v2 \citep{chen2023minigpt}, and Fuyu-8B \citep{Fuyu-8B}. We primarily focus on the occurrence of object hallucination within the generated descriptions. To this end, we randomly select 5,000 images from the MSCOCO validation set \citep{lin2014microsoft}  and prompt the LVLMs to generate detailed descriptions of these images. Then we employ the CHAIR evaluation framework \citep{rohrbach2018object} for our analysis. The corresponding metrics are formulated as follows:
\begin{equation}
\small
C_s=\frac{|\{\text{hallucinated objects}\}|}{|\{\text{all mentioned objects}\}|}, C_i=\frac{|\{\text{captions with hallucinated objects}\}|}{|\{\text{all captions}\}|}.
\end{equation} 
Higher $C_s$ and $C_i$ indicate more serious hallucinations. In terms of decoding strategy selection, we adopt two commonly used decoding strategies, including greedy decoding and random multinomial sampling decoding \citep{wolf2019huggingface}. Since the proposed MiHO and MiHI are two orthogonal methods, we report the results of MiHO, MiHI, and the combined use of MiHO and MiHI.

\textbf{Q1 Hypothesis verification.} In Table~\ref{table:q1}, we report the hallucination evaluation performance for content before `\textbackslash n' compared to content generated after `\textbackslash n'. We can see that the content produced after `\textbackslash n' has a significant probability of hallucination. 

\begin{table}[!htbp]
\scriptsize
  \centering
  \caption{Q1 Hypothesis verification. Sentences generated after `\textbackslash n' have more hallucinations.}
    \begin{tabular}{c|cc|cc|cc|cc|cc|cc}
    \toprule
    Model & \multicolumn{4}{c|}{BakLLaVA  \citep{liu2023improved}}  & \multicolumn{4}{c|}{InstructBLIP-7B \citep{li2023blip}} & \multicolumn{4}{c}{LLaVA-v1.5-7B \citep{liu2023improved}} \\
    Decoding & \multicolumn{2}{c|}{Greedy} & \multicolumn{2}{c|}{Sampling} & \multicolumn{2}{c|}{Greedy} & \multicolumn{2}{c|}{Sampling} & \multicolumn{2}{c|}{Greedy} & \multicolumn{2}{c}{Sampling} \\
    Method & $C_s\downarrow$ & $C_i\downarrow$ & $C_s\downarrow$ & $C_i\downarrow$ & $C_s\downarrow$ & $C_i\downarrow$ & $C_s\downarrow$ & $C_i\downarrow$ & $C_s\downarrow$ & $C_i\downarrow$ & $C_s\downarrow$ & $C_i\downarrow$ \\
    \midrule
    Before \textbackslash n & 18.14 & 6.09  & 22.28 & 7.83  & 48.12 & 14.47 & 55.96 & 17.66 & 21.24 & 7.37  & 29.98 & 11.30 \\
    After \textbackslash n & 53.42 & 23.33 & 53.84 & 24.46 & 55.88 & 29.45 & 63.29 & 33.33 & 57.23 & 28.06 & 59.99 & 30.62 \\
    \midrule
    Model & \multicolumn{4}{c|}{Fuyu-8B \citep{Fuyu-8B}}   & \multicolumn{4}{c|}{MiniGPT-v2 \citep{chen2023minigpt}} & \multicolumn{4}{c}{LLaVA-v1.5-13B \citep{liu2023improved}} \\
    Decoding & \multicolumn{2}{c|}{Greedy} & \multicolumn{2}{c|}{Sampling} & \multicolumn{2}{c|}{Greedy} & \multicolumn{2}{c|}{Sampling} & \multicolumn{2}{c|}{Greedy} & \multicolumn{2}{c}{Sampling} \\
    Method & $C_s\downarrow$ & $C_i\downarrow$ & $C_s\downarrow$ & $C_i\downarrow$ & $C_s\downarrow$ & $C_i\downarrow$ & $C_s\downarrow$ & $C_i\downarrow$ & $C_s\downarrow$ & $C_i\downarrow$ & $C_s\downarrow$ & $C_i\downarrow$ \\
    \midrule
    Before \textbackslash n & 17.36 & 6.84  & 27.04 & 10.31 & 24.26 & 8.91  & 30.92 & 11.95 & 20.12 & 6.60  & 26.72 & 9.20 \\
    After \textbackslash n & 58.22 & 29.23 & 61.53 & 30.16 & 58.83 & 31.49 & 62.09 & 32.40 & 52.11 & 23.77 & 57.66 & 26.93 \\
    \bottomrule
    \end{tabular}%
\label{table:q1}
\end{table}%
% TODO 加一个直观一些的图

% TODO USER的prompt长度对\n\n的影响
% use case: continue, user first image ask llm to continue des
% 比一些baseline的方法
\textbf{Q2 Attackability.} We also validate that inserting  `\textbackslash n\textbackslash n' at appropriate positions can induce LVLMs to generate more hallucinations. Specifically, when the sentence outputs the period (`.') token for the $k$-th time, we manually insert the `\textbackslash n\textbackslash n' to initiate the attack, where $k$ is a manually specified position hyperparameter. The experimental results are shown in Table~\ref{table:q3}. The table indicates that inserting `\textbackslash n\textbackslash n' later in the sentence increases hallucinations.
% Table generated by Excel2LaTeX from sheet '表格2 重做'
\begin{table}[htbp]
  \centering
  \scriptsize
  \caption{Q2 Attackability. Performance comparison when attack LVLMs at different position, where Attack-$k$ refers to initiating the attack upon encountering the period (`.') token for the $k$-th time.}
    \begin{tabular}{c|cc|cc|cc|cc|cc|cc}
    \toprule
    Model & \multicolumn{4}{c|}{BakLLaVA  \citep{liu2023improved}}  & \multicolumn{4}{c|}{InstructBLIP-7B \citep{li2023blip}} & \multicolumn{4}{c}{LLaVA-v1.5-7B \citep{liu2023improved}} \\
    Decoding & \multicolumn{2}{c|}{Greedy} & \multicolumn{2}{c|}{Sampling} & \multicolumn{2}{c|}{Greedy} & \multicolumn{2}{c|}{Sampling} & \multicolumn{2}{c|}{Greedy} & \multicolumn{2}{c}{Sampling} \\
    Method & $C_s\downarrow$    & $C_i\downarrow$    & $C_s\downarrow$    & $C_i\downarrow$    & $C_s\downarrow$    & $C_i\downarrow$    & $C_s\downarrow$    & $C_i\downarrow$    & $C_s\downarrow$    & $C_i\downarrow$    & $C_s\downarrow$    & $C_i\downarrow$ \\
    \midrule
    Original     & 48.56 & 13.00 & 51.58 & 14.37 & 48.44 & 14.53 & 56.66 & 17.81 & 51.74 & 15.35 & 57.16 & 18.47 \\
    Attack-$1$ & 47.42 & 12.64 & 51.72 & 14.33 & 58.28 & 17.29 & 63.76 & 19.80 & 55.28 & 16.42 & 60.60 & 19.90 \\
    Attack-$2$ & 49.36 & 13.23 & 55.16 & 15.55 & 59.52 & 18.63 & 65.74 & 20.81 & 56.42 & 16.72 & 62.48 & 20.46 \\
    Attack-$3$ & 54.54 & 13.83 & 58.28 & 15.56 & 58.18 & 18.13 & 65.76 & 20.69 & 60.58 & 16.95 & 65.96 & 20.71 \\
    \midrule
    Model & \multicolumn{4}{c|}{Fuyu-8B \citep{Fuyu-8B}}   & \multicolumn{4}{c|}{MiniGPT-v2 \citep{chen2023minigpt}} & \multicolumn{4}{c}{LLaVA-v1.5-13B \citep{liu2023improved}} \\
    Decoding & \multicolumn{2}{c|}{Greedy} & \multicolumn{2}{c|}{Sampling} & \multicolumn{2}{c|}{Greedy} & \multicolumn{2}{c|}{Sampling} & \multicolumn{2}{c|}{Greedy} & \multicolumn{2}{c}{Sampling} \\
    Method & $C_s\downarrow$    & $C_i\downarrow$    & $C_s\downarrow$    & $C_i\downarrow$    & $C_s\downarrow$    & $C_i\downarrow$    & $C_s\downarrow$    & $C_i\downarrow$    & $C_s\downarrow$    & $C_i\downarrow$    & $C_s\downarrow$    & $C_i\downarrow$ \\
    \midrule
    Original     & 59.74 & 17.27 & 63.08 & 19.18 & 62.62 & 19.50 & 64.36 & 21.08 & 47.52 & 13.08 & 53.04 & 15.69 \\
    Attack-1 & 51.26 & 15.35 & 59.28 & 18.48 & 61.26 & 18.31 & 64.14 & 20.88 & 47.56 & 13.08 & 53.16 & 15.96 \\
    Attack-2 & 54.60 & 17.16 & 63.18 & 19.80 & 63.16 & 20.30 & 67.78 & 22.47 & 50.50 & 14.27 & 57.32 & 17.51 \\
    Attack-3 & 63.14 & 17.98 & 68.74 & 20.69 & 65.50 & 20.17 & 71.64 & 22.79 & 54.62 & 14.15 & 61.12 & 17.08 \\
    \bottomrule
    \end{tabular}%
  \label{table:q3}%
\end{table}%

\textbf{Q3 Effectiveness.} As shown in Table~\ref{table:q2_1}, we compare the proposed methods with the original outputs of LVLMs. Furthermore, to eliminate the influence of sentence length on the output, we conduct a comparison between the proposed and original methods at equal output lengths, achieved by truncating the end of each sentence. The experimental results are shown in Table~\ref{table:q2_2}. Based on these experimental results, we can draw the following conclusions: (1) As shown in Table~\ref{table:q2_1}, both MiHO and MiHO+MiHI significantly reduce the occurrence of hallucinations across all models. MiHI also significantly reduces hallucinations in all models except Fuyu-8B, possibly because Fuyu-8B is not fine-tuned with instructions, resulting in a poorer understanding of prompts. (2) According to Table~\ref{table:q2_2}, when comparing original descriptions of the same length, MiHO demonstrates significant improvements in almost all models. However, MiHI and MiHO+MiHI sometimes exhibit performance decreases, possibly because the modified prompts negatively impacts the descriptions of LVLMs. (3) Compared to greedy decoding, sampling decoding strategy is more prone to producing hallucinations. Our method shows better performance when used with greedy decoding strategy. 

% Table generated by Excel2LaTeX from sheet '表格1'
\begin{table}[!htbp]
  \centering
  \caption{Q3 Effectiveness. Performance comparison of the proposed method on different LVLMs.}
  \scriptsize
    \begin{tabular}{c|cc|cc|cc|cc|cc|cc}
    \toprule
    Model & \multicolumn{4}{c|}{BakLLaVA \citep{liu2023improved}}  & \multicolumn{4}{c|}{InstructBLIP-7B \citep{li2023blip}} & \multicolumn{4}{c}{LLaVA-v1.5-7B \citep{liu2023improved}} \\
    Decoding & \multicolumn{2}{c|}{Greedy} & \multicolumn{2}{c|}{Sampling} & \multicolumn{2}{c|}{Greedy} & \multicolumn{2}{c|}{Sampling} & \multicolumn{2}{c|}{Greedy} & \multicolumn{2}{c}{Sampling} \\
    Method & $C_s\downarrow$    & $C_i\downarrow$    & $C_s\downarrow$    & $C_i\downarrow$    & $C_s\downarrow$    & $C_i\downarrow$    & $C_s\downarrow$    & $C_i\downarrow$    & $C_s\downarrow$    & $C_i\downarrow$    & $C_s\downarrow$    & $C_i\downarrow$ \\
    \midrule
    Original     & 48.56 & 13.00 & 51.58 & 14.37 & 48.44 & 14.53 & 56.66 & 17.81 & 51.74 & 15.35 & 57.16 & 18.47 \\
    \midrule
    MiHO     & 38.96 & 10.35 & 42.66 & 11.87 & 48.30 & 14.50 & 57.06 & 18.27 & 38.62 & 11.35 & 47.70 & 15.35 \\
    MiHI     & 42.04 & 11.66 & 47.10 & 13.39 & 45.70 & 12.91 & 56.26 & 17.11 & 39.40 & 12.54 & 45.32 & 16.02 \\
    MiHO+MiHI  & 36.68 & 10.04 & 42.16 & 11.93 & 45.70 & 12.91 & 57.40 & 17.34 & 39.38 & 12.53 & 45.36 & 16.03 \\
    \midrule
    Model & \multicolumn{4}{c|}{Fuyu-8B \citep{Fuyu-8B}}  & \multicolumn{4}{c|}{MiniGPT-v2 \citep{chen2023minigpt}}  & \multicolumn{4}{c}{LLaVA-v1.5-13B \citep{liu2023improved}} \\
    Decoding & \multicolumn{2}{c|}{Greedy} & \multicolumn{2}{c|}{Sampling} & \multicolumn{2}{c|}{Greedy} & \multicolumn{2}{c|}{Sampling} & \multicolumn{2}{c|}{Greedy} & \multicolumn{2}{c}{Sampling} \\
    % \hline
    Method & $C_s\downarrow$    & $C_i\downarrow$    & $C_s\downarrow$    & $C_i\downarrow$    & $C_s\downarrow$    & $C_i\downarrow$    & $C_s\downarrow$    & $C_i\downarrow$    & $C_s\downarrow$    & $C_i\downarrow$    & $C_s\downarrow$    & $C_i\downarrow$ \\
    \midrule
    Original     & 59.74 & 17.27 & 63.08 & 19.18 & 62.62 & 19.50 & 64.36 & 21.08 & 47.52 & 13.08 & 53.04 & 15.69 \\
    \midrule
    MiHO     & 38.14 & 10.60 & 45.92 & 14.44 & 33.02 & 11.38 & 45.66 & 15.60 & 35.32 & 9.61  & 43.58 & 13.15 \\
    MiHI     & 59.08 & 17.37 & 63.40 & 19.54 & 49.28 & 14.99 & 58.16 & 19.13 & 37.64 & 10.54 & 47.26 & 14.25 \\
    MiHO+MiHI  & 40.76 & 11.16 & 48.36 & 14.94 & 42.38 & 13.24 & 51.34 & 17.62 & 34.64 & 9.73  & 44.72 & 13.45 \\
    \bottomrule
    \end{tabular}%
\label{table:q2_1}
\end{table}%

% Table generated by Excel2LaTeX from sheet '表格3重做'
\begin{table}[htbp]
\scriptsize
  \centering
  \caption{Q3 Effectiveness. Performance comparison of the proposed method on different LVLMs when the output sentence lengths are equal.}
    \begin{tabular}{c|cc|cc|cc|cc|cc|cc}
    \toprule
    Model & \multicolumn{4}{c|}{BakLLaVA  \citep{liu2023improved}}  & \multicolumn{4}{c|}{InstructBLIP-7B \citep{li2023blip}} & \multicolumn{4}{c}{LLaVA-v1.5-7B \citep{liu2023improved}} \\
    Decoding & \multicolumn{2}{c|}{Greedy} & \multicolumn{2}{c|}{Sampling} & \multicolumn{2}{c|}{Greedy} & \multicolumn{2}{c|}{Sampling} & \multicolumn{2}{c|}{Greedy} & \multicolumn{2}{c}{Sampling} \\
    Method & $C_s\downarrow$    & $C_i\downarrow$    & $C_s\downarrow$    & $C_i\downarrow$    & $C_s\downarrow$    & $C_i\downarrow$    & $C_s\downarrow$    & $C_i\downarrow$    & $C_s\downarrow$    & $C_i\downarrow$    & $C_s\downarrow$    & $C_i\downarrow$ \\
    \midrule
    Original & 44.00 & 11.93 & 45.78 & 13.11 & 48.38 & 14.49 & 52.52 & 16.71 & 45.04 & 13.24 & 49.40 & 15.98 \\
    MiHO     & 38.96 & 10.35 & 42.66 & 11.87 & 48.30 & 14.50 & 53.08 & 17.21 & 38.62 & 11.35 & 47.70 & 15.35 \\
    \midrule
    Original & 44.34 & 12.20 & 45.88 & 13.20 & 45.70 & 13.35 & 51.82 & 16.55 & 36.16 & 10.98 & 41.82 & 14.26 \\
    MiHI     & 42.04 & 11.66 & 47.10 & 13.39 & 45.70 & 12.91 & 52.18 & 16.24 & 39.40 & 12.54 & 45.32 & 16.02 \\
    \midrule
    Original & 41.64 & 11.44 & 43.30 & 12.55 & 45.70 & 13.35 & 52.52 & 16.71 & 36.16 & 10.98 & 41.58 & 14.32 \\
    MiHO+MiHI   & 36.68 & 10.04 & 42.16 & 11.93 & 45.70 & 12.91 & 53.66 & 16.49 & 39.38 & 12.53 & 45.36 & 16.03 \\
    \midrule
    Model & \multicolumn{4}{c|}{Fuyu-8B \citep{Fuyu-8B}}   & \multicolumn{4}{c|}{MiniGPT-v2 \citep{chen2023minigpt}} & \multicolumn{4}{c}{LLaVA-v1.5-13B \citep{liu2023improved}} \\
    Decoding & \multicolumn{2}{c|}{Greedy} & \multicolumn{2}{c|}{Sampling} & \multicolumn{2}{c|}{Greedy} & \multicolumn{2}{c|}{Sampling} & \multicolumn{2}{c|}{Greedy} & \multicolumn{2}{c}{Sampling} \\
    Method & $C_s\downarrow$    & $C_i\downarrow$    & $C_s\downarrow$    & $C_i\downarrow$    & $C_s\downarrow$    & $C_i\downarrow$    & $C_s\downarrow$    & $C_i\downarrow$    & $C_s\downarrow$    & $C_i\downarrow$    & $C_s\downarrow$    & $C_i\downarrow$ \\
    \midrule
    Original & 44.36 & 13.10 & 51.06 & 16.00 & 35.04 & 12.41 & 49.34 & 17.12 & 42.10 & 11.49 & 46.50 & 13.95 \\
    MiHO     & 38.14 & 10.60 & 45.92 & 14.44 & 33.02 & 11.38 & 45.66 & 15.60 & 35.32 & 9.61  & 43.58 & 13.15 \\
    \midrule
    Original & 58.30 & 16.90 & 59.08 & 18.00 & 54.90 & 16.77 & 56.90 & 18.86 & 42.42 & 11.57 & 47.02 & 14.12 \\
    MiHI     & 59.08 & 17.37 & 63.40 & 19.54 & 49.28 & 14.99 & 58.16 & 19.13 & 37.64 & 10.54 & 47.26 & 14.25 \\
    \midrule
    Original & 47.62 & 13.90 & 53.00 & 16.29 & 49.44 & 15.59 & 52.56 & 17.58 & 41.04 & 11.30 & 45.80 & 13.76 \\
    MiHO+MiHI    & 40.76 & 11.16 & 48.36 & 14.94 & 42.38 & 13.24 & 51.34 & 17.62 & 34.64 & 9.73  & 44.72 & 13.45 \\
    \bottomrule
    \end{tabular}%
  \label{table:q2_2}%
\end{table}%

\section{Related Work}
\textbf{Hallucination mitigation in LVLMs} can be primarily categorized into two types including retraining-based methods and post-hoc processing-based methods. Retraining-based methods include collecting high-quality data \citep{wang2023vigc}, modifying the architecture of LVLMs \citep{tong2024eyes,zhai2023halle}, and adjusting the training strategy for LVLMs \citep{yu2023rlhf,zhao2023beyond,sun2023aligning,ben2023mocha}. Although these methods are effective, they often require additional computational overhead to alleviate hallucinations of LVLMs. On the other hand, post-hoc processing methods aim to mitigate hallucination without retraining the LVLMs. These methods involve modifying the decoding method of LVLMs \citep{huang2023opera,leng2023mitigating}, enriching the visual context of LVLMs by integrating existing open-source vision models \citep{zhao2024mitigating}, and training an additional reviser model \citep{zhou2023analyzing}.

\textbf{Bias in machine learning.} Bias in machine learning models, caused by biased training data, can lead to distrust and serious consequences \citep{elazar2023s,longpre2023pretrainer}. It has been widely explored include bias against minority subpopulations \citep{han2022umix,han2023reweighted}, bias against datasets \citep{torralba2011unbiased,liu2024decade}, bias in alignment of image and text information \cite{tong2024eyes, lin2023parrot}. In this paper we focus on the semantic shift bias in LVLMs triggered by paragraph breaks.

\section{Conclusions and Takeaways}
In this paper, we identify a phenomenon of `\textbackslash n\textbackslash n'-induced hallucinations in some existing LVLMs, attributed to semantic shift bias. Besides, we find that inserting `\textbackslash n\textbackslash n' during the description generation process can induce hallucinations in LVLMs. Based on this, we propose a method to alleviate hallucinations by reducing the probability of `\textbackslash n' from the input and output perspectives. Extensive experiments on multiple publicly available LVLMs are conducted  to verify the performance of our method. It should be highlighted that the `\textbackslash n\textbackslash n'-induced hallucination problem is not found in some LVLMs, \textit{e.g.}, GPT-4 \citep{achiam2023gpt}. What causes the hallucination problem remains an open question. Finally, it remains to be explored whether this bias can be overcome when the model scale continues to increase.

\section{Acknowledgement}
This work is supported in part by the scholarship from China Scholarship Council (No.202306250107). We also appreciate discussions with Shiwei Wu and his suggestions.
% This could potentially be attributed to the relative frequency of `\textbackslash n\textbackslash n' in the instruction tuning stage. Notably, in the GPT-4 generated instruction tuning dataset, a substantial occurrence of `\textbackslash n\textbackslash n' is observed. This finding indicates that high-quality instruction datasets or more reasonable instruction tuning methods to avoid introducing the above data bias may be necessary.

\bibliography{iclr2024_conference}
\bibliographystyle{iclr2024_conference}
\newpage
\appendix
% Table generated by Excel2LaTeX from sheet 'BeamSearch表格4-与Opera对比'

\section{More Experimental Results}
In this section, we present further experimental evidence to validate the effectiveness of our proposed methodology. We conduct comparisons with two contemporary approaches: DoLa \citep{chuang2023dola} and OPERA \citep{huang2023opera}. Given that both referenced methods employ the Beam Search decoding strategy, we also incorporate the Beam Search strategy into our proposed method to ensure a fair comparison. The results of these experiments are shown in Table \ref{tab:table5}. The experimental results demonstrate that our proposed method achieves the best performance.

\begin{table}[!htbp]
  \centering
  \caption{Comparison with current state-of-the-art methods. The experimental results of Beam Search, DoLa and OPERA are from OPERA \citep{huang2023opera}.}
    \begin{tabular}{c|cc|cc}
    \toprule
    Model & \multicolumn{2}{c|}{Llava-1.5-7B} & \multicolumn{2}{c}{InstructBLIP} \\
    % \midrule
    Method & $C_s$ & $C_i$    & $C_s$ & $C_i$ \\
    \midrule
    Beam Search \citep{graves2012sequence} & 48.8  & 13.9  & 55.6  & 15.8 \\
    DoLa \citep{chuang2023dola}  & 47.8  & 13.8  & 48.4  & 15.9 \\
    OPERA \citep{huang2023opera} & 44.6  & 12.8  & 46.4  & 14.2 \\
    \midrule
    MiHO  & \textcolor{brown}{\textbf{34.6}}  & \textcolor{brown}{\textbf{10.2}}  & 47.9  & 13.8 \\
    MiHI  & 37.1  & 11.4  & 44.7  & 11.7 \\
    MiHI+MiHO & 37.1  & 11.4  & \textcolor{brown}{\textbf{44.7}}  & \textcolor{brown}{\textbf{11.7}} \\
    \bottomrule
    \end{tabular}%
  \label{tab:table5}%
\end{table}%

To further analyze the impact of the prompt in MiHI, we conducted additional experiments with different prompts generated by GPT-4. The prompts are shown as follows.
\begin{enumerate}
\item[\textbf{P1}] \emph{Please describe this image in detail \textcolor{brown}{in one paragraph}.''} 
\item[\textbf{P2}]  \emph{Please describe this image in detail \textcolor{brown}{in a single, continuous text}.''}
\item[\textbf{P3}]  \emph{``Please describe this image in detail \textcolor{brown}{, with no separation into paragraphs}.''}
\item[\textbf{P4}]  \emph{Please describe this 
image in detail \textcolor{brown}{without \textbackslash n}.''}
\item[\textbf{P5}]  \emph{Please describe this image in detail \textcolor{brown}{without using paragraph breaks}.''} 
\end{enumerate}
The experimental results are shown in Table~\ref{tab:moreprompt}. It can be seen from the experimental results that prompts may have a significant impact on hallucinations.

% Table generated by Excel2LaTeX from sheet '更多的prompt2'
\begin{table}[!htbp]
  \centering
   \scriptsize 
  \caption{Results of the proposed method using various prompts.}
    \begin{tabular}{c|cc|cc|cc|cc|cc|cc}
    \toprule
    Model & \multicolumn{2}{c|}{BakLLaVA} & \multicolumn{2}{c|}{InstructBLIP-7B} & \multicolumn{2}{c|}{LLaVA-v1.5-7B} & \multicolumn{2}{c|}{Fuyu-8B} & \multicolumn{2}{c|}{MiniGPT-v2} & \multicolumn{2}{c}{LLaVA-1.5-13B} \\
    \midrule
    Method & $C_s$ & $C_i$    & $C_s$ & $C_i$    & $C_s$ & $C_i$    & $C_s$ & $C_i$    & $C_s$ & $C_i$    & $C_s$ & $C_i$ \\
    \midrule
    P1    & 42.04 & 11.66 & 45.70 & 12.91 & 39.40 & 12.54 & 59.08 & 17.37 & 49.28 & 14.99 & 37.64 & 10.54 \\
    P2    & 47.44 & 12.71 & 30.46 & 9.88  & 45.08 & 12.62 & 61.66 & 18.30 & 56.08 & 18.03 & 49.24 & 14.62 \\
    P3    & 48.02 & 12.74 & 40.06 & 11.40 & 46.62 & 13.03 & 58.16 & 17.24 & 55.78 & 18.17 & 50.86 & 15.16 \\
    P4    & 46.24 & 12.38 & 48.86 & 14.88 & 44.28 & 12.40 & 59.72 & 17.95 & 62.70 & 19.76 & 48.82 & 14.58 \\
    P5    & 47.22 & 12.58 & 45.00 & 12.93 & 46.08 & 12.85 & 57.70 & 16.76 & 55.96 & 18.38 & 51.02 & 15.26 \\
    \midrule
    P1+MiHO & 36.68 & 10.04 & 45.70 & 12.91 & 39.38 & 12.53 & 40.76 & 11.16 & 42.38 & 13.24 & 34.64 & 9.73 \\
    P2+MiHO & 38.76 & 10.25 & 30.46 & 9.88  & 34.84 & 9.53  & 40.26 & 11.08 & 32.76 & 11.64 & 40.08 & 11.71 \\
    P3+MiHO & 39.46 & 10.39 & 39.98 & 11.38 & 35.54 & 9.66  & 38.24 & 10.71 & 33.62 & 11.82 & 39.86 & 11.73 \\
    P4+MiHO & 38.50 & 10.22 & 48.40 & 14.74 & 34.92 & 9.65  & 38.60 & 10.50 & 31.72 & 11.33 & 38.04 & 11.38 \\
    P5+MiHO & 39.16 & 10.53 & 44.96 & 12.92 & 36.44 & 9.90  & 38.90 & 10.72 & 30.26 & 11.26 & 40.90 & 11.99 \\
    \bottomrule
    \end{tabular}%
  \label{tab:moreprompt}%
\end{table}%

\end{document}